\pdfoutput=1

\documentclass[11pt]{article}

\usepackage[]{acl}

\usepackage{times}
\usepackage{latexsym}

\usepackage[T1]{fontenc}

\usepackage[utf8]{inputenc}

\usepackage{microtype}

\usepackage{booktabs}
\usepackage{multirow}
\usepackage{graphicx}
\usepackage{arydshln}

\title{Solving NLP Problems through Human-System Collaboration: A Discussion-based Approach}

\author{Masahiro Kaneko$^{1}$ \quad
        Graham Neubig$^{2}$\quad
        Naoaki Okazaki$^{1}$ \\
        $^1$Tokyo Institute of Technology \quad
        $^2$Carnegie Mellon University \\
        {\tt masahiro.kaneko@nlp.c.titech.ac.jp} \\
        {\tt gneubig@cs.cmu.edu} \quad
        {\tt okazaki@c.titech.ac.jp}
}

\begin{document}
\maketitle
\begin{abstract}

Humans work together to solve common problems by having discussions, explaining, and agreeing or disagreeing with each other.
Similarly, if a system can have discussions with human partners when solving tasks, it has the potential to improve the system's performance and reliability.
In previous research on explainability, it has only been possible for systems to make predictions and for humans to ask questions about them, rather than having a mutual exchange of opinions.
This research aims to create a dataset\footnote{Our dataset is publicly available at:\url{https://github.com/kanekomasahiro/discussion_nlp}} and a computational framework for systems that discuss and refine their predictions through dialogue. 
Through experiments, we show that the proposed system can have beneficial discussions with humans, improving the accuracy by up to 25 points on a natural language inference task.

\end{abstract}

\section{Introduction}
\label{sec:intro}

Today's deep learning systems are performant but opaque, leading to a wide variety of explainability techniques that attempt to take in a system prediction and output an explanation justifying the prediction~\cite{ribeiro2016should,shwartz2017opening,fong2017interpretable,kim2018interpretability,lipton2018mythos,wiegreffe-etal-2022-reframing}.
Many such explainability techniques require significant expertise in deep learning to use effectively, requiring consumers of the explanations to analyze the data, internal states, and output trends of the system of interest~\cite{ribeiro2016should,kaneko-etal-2022-interpretability,Kaneko2023ControlledGW}.
However, many potential system users lack this expertise, such as medical or legal professionals who want to use machine learning models and need to confirm the veracity of the generated results or rectify any mistaken predictions.

To address this issue, researchers are working to find ways to both explain system predictions in natural language~\cite{Ling2017-xk,Raffel2019-nq,Brown2020-eh,wiegreffe-etal-2022-reframing,du2023improving} and give instructions and feedback to systems through natural language~\cite{abramson2022improving,sharma2022correcting,murty2022fixing,campos2022training,bowman2022measuring,Loem2023SAIEFS}.
Chain-of-Thought (CoT) prompting has shown that natural language contributes to performance improvements in complex multistep inference~\cite{wei2022chain,wang2022self,zhang2022automatic}.
Step-by-step reasoning in CoT relies solely on the system to make predictions without human involvement.
There is also work that allows users to ask questions about the system's predictions and tasks~\cite{Slack2022-kh} in a conversational format.
Compared to the more standard learning and explanation paradigms, this approach allows humans to understand and teach the system intuitively.
However, in these works, the communication tends to be one-sided, from human-to-system or system-to-human, which still falls short of the full interactive problem solving process experienced by human interlocutors~\cite{Lakkaraju2022-gt}.

\begin{figure}[t!]
\centering
\includegraphics[width=8.cm]{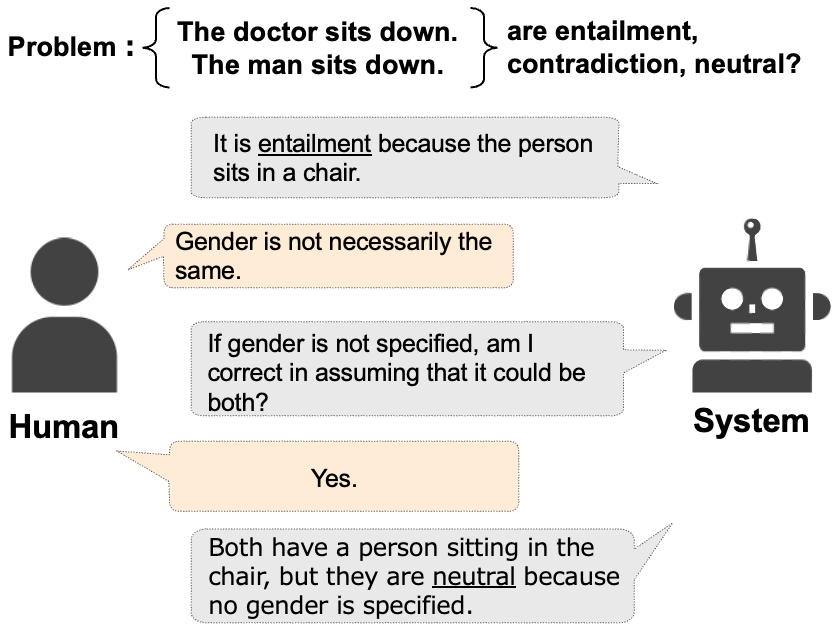}
\caption{Human-system discussions in NLI.}
\label{fig:abst}
\end{figure}

In this study, we take the first steps towards establishing a framework for \emph{human-system collaboration on prediction problems through discussion} (illustration in \autoref{fig:abst}).
If such a system is realized, it will allow both humans and the system to engage in explanations of predictions, ask questions about unclear points, refine their thoughts, and solve problems.

First, we create a dataset of \emph{human-human discussions} regarding a prediction task (Section \ref{sec:dataset_creation}).
In particular, we use the task of natural language inference (NLI): prediction of the relationship between a ``premise'' sentence and a ``hypothesis'' sentence is entailment, contradiction, or neutral~\cite{Bowman2015-ib}.
We specifically choose relatively \textit{difficult or ambiguous cases} to spur discussion between the participants.

Second, we train and evaluate a system that is capable of discussing an NLI problem with a human (Sections \ref{sec:discussion_system}, \ref{sec:evaluation_method}). 
It is achieved by constructing prompts with manually created discussion examples so the system can learn from humans how to discuss, accept, or object to the provided opinions about the topic.

The results of both quantitative and human evaluation demonstrate that a system could perform more informative discussions by training to have a discussion with few-shot learning (Section \ref{sec:experiments}).
We also found that providing the system with information about the discussion topic improved its performance in many cases compared to the system that did not have access to such information.
On the other hand, the discussion revealed that the system tends to be too compliant with human opinions. 
Therefore, addressing the risk of transmitting incorrect knowledge or maliciously altering the system's knowledge of humans is necessary.
We also show that few-shot usage of discussion data can enable the system to counter human arguments correctly (Section \ref{sec:analysis}).
Finally, we demonstrate that using discussion data generated by the system~\cite{wang2022self,huang2022large} can achieve equivalent results to those of the system that used manually created discussion data in few-shot learning or fine-tuning cases.

\begin{figure*}[t!]
\centering
\includegraphics[width=16cm]{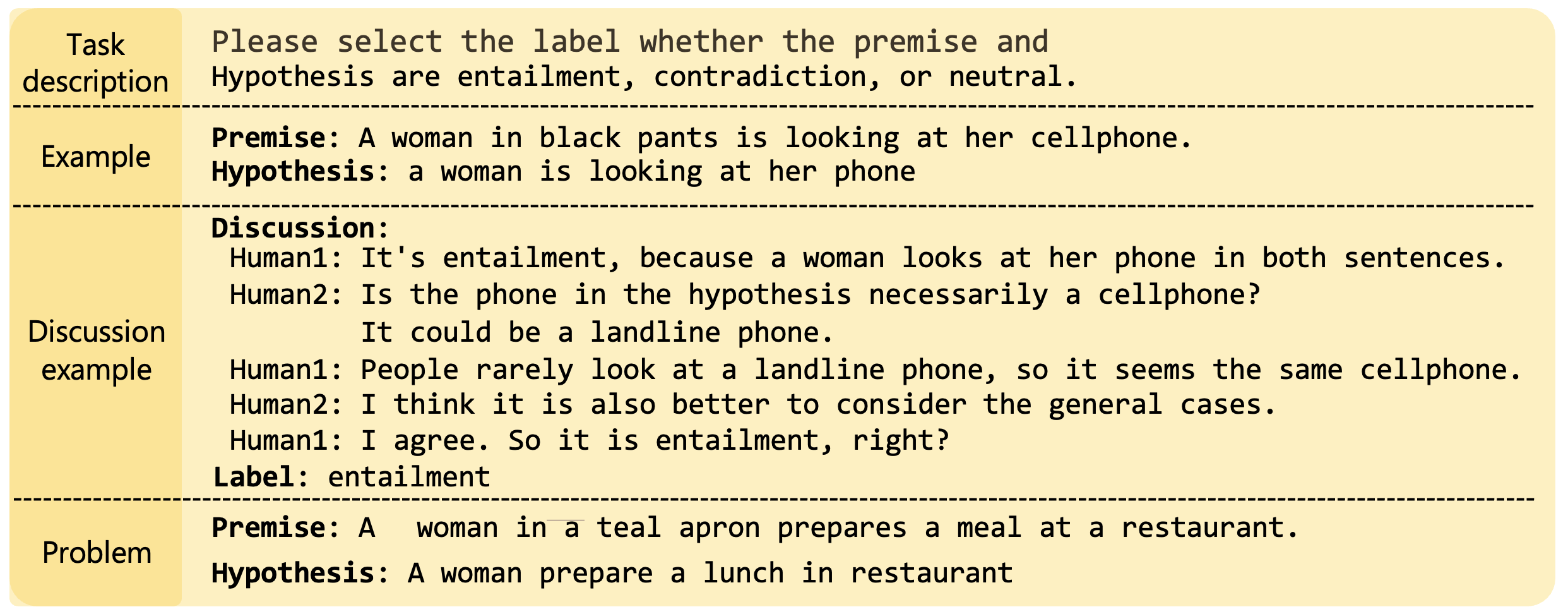}
\caption{Prompt with a single example for few-shot learning.}
\label{fig:prompt}
\end{figure*}

\section{Discussion Dataset Creation}
\label{sec:dataset_creation}


The NLI task aims to determine the logical relationship between a hypothesis sentence and a premise sentence~\cite{Bowman2015-ib}.
The task involves classifying whether the hypothesis sentence is entailment, contradiction, or neutral.
For example, given the premise \textit{``The cat is sitting on the mat''} and the hypothesis \textit{"The mat is empty"}, the task would involve classifying the relationship as a contradiction.
NLI tasks require deep assimilation of fine nuances of common sense knowledge, and much work has been done to explain this with natural language as a prediction reason~\cite{Camburu2018-ho,kumar-talukdar-2020-nile}.
Therefore, we also target the NLI task and build a system that predicts entailment, contradiction, or neutrality through discussion.

To train a system that can engage in a discussion, we create a dataset of human annotators discussing NLI problems.
We use the Stanford NLI (SNLI) dataset~\cite{Bowman2015-ib}, a common benchmark dataset in NLP, to create the discussion data.
Collecting high-quality discussion data among humans is costly, as it requires knowledgeable annotators about the task and multiple dialogue turns for each problem.
Fourteen annotators with knowledge of NLP were asked to annotate the data.\footnote{Annotation work was requested at \$25 per hour.
The data collection from human participants was conducted under an institutional review board protocol.}

First, the annotators were presented with premise and hypothesis sentences and asked to predict labels such as entailment, contradiction, or neutral.
We randomly paired two annotators to have them assign labels for the same premise and hypothesis.
Then, they discussed the labels that they had assigned differently and decided on the final labels based on those discussions.
The premise and hypothesis sentences were sampled from 300 problems from the development data and 750 problems from the evaluation data of SNLI.
These were used as development and evaluation data in the discussion data, respectively.
Each annotator pair is asked to predict the labels of 150 problems.
SNLI development data originally consists of problems with labels from five crowd workers, and the majority vote of these labels determines the golden label.
To find relatively hard cases that might spur more discussion, we sampled problems for annotation from those in which three of the five had the same label.

Our annotators were then paired with each other and discussed the questions for which they had given different labels.
They discussed in a free-form manner until they agreed on a final decision.%
\footnote{They were also instructed not to include personal information and inappropriate utterances.}
Preliminary experimental results showed that the number of discussion turns tended to be higher for oral rather than text-based discussions.
Therefore, we created discussion data by transcribing oral discussions among the annotators, using Whisper (medium.en)~\cite{radford2022robust}\footnote{\url{https://github.com/openai/whisper}} for transcription.
The text transcribed by Whisper was manually corrected for transcription errors and manually separated into speech segments.

Then, for each utterance, we assigned the evidential utterances for the final label and the labels of ``\textit{supportive}'', ``\textit{unsupportive}'', or ``\textit{irrelevant}'' to each utterance.
For example, for \autoref{fig:abst}, ``\textit{Both have a person sitting in the chair, but they are neutral because no gender is specified.}'' is labeled as supportive, ``\textit{It is entailment because the person sits in a chair.}'' is unsupportive, and ``\textit{Yes.}'' is labeled as irrelevant.
These labels are not used in the few-shot learning process but are used to \textit{evaluate the discussion ability of the system} automatically.

In this annotation work, discussion data were collected for 102 problems.
Of these, 10 problems were used as prompts for few-shot learning, 27 for validation data, and 65 for evaluation data.
The average number of utterances for each problem in the prompt, validation, and evaluation data is 4.4, 6.3, and 5.1 respectively.
For validation and evaluation data, the number of supportive/unsupportive utterances are 85/23 and 133/72 respectively.

\section{Discussion System}
\label{sec:discussion_system}

We use three types of systems in the experiments: \textbf{zero-shot}, \textbf{few-shot}, and \textbf{few-shot-discussion}.
In the zero-shot system, only the task description is given as a prompt.
In the few-shot system, the examples' task description and premise, hypothesis, and gold labels are given as prompts.
In the few-shot-discussion system, in addition to the task description and examples, human discussion examples about the labels of the examples are given as prompts.
These prompts are concatenated with the problem to be solved and given as input to the system to perform inference.
Examples of each prompt are shown in \autoref{fig:prompt}.
The discussion example distinguishes human utterances between ``\textit{Human1:}'' and ``\textit{Human2:}''.

The examples used in the prompts are the same for both the few-shot and the few-shot-discussion systems.
We use the same examples for all problems.
All methods do not update the parameters of the systems.
We use GPT-3.5\footnote{text-davinci-003: \url{https://beta.openai.com/docs/models/gpt-3}}~\cite{Brown2020-eh} and ChatGPT\footnote{gpt-3.5-turbo: \url{https://platform.openai.com/docs/guides/gpt/chat-completions-api}}~\cite{chatgpt} for the zero-shot, few-shot, and few-shot-discussion systems.

\section{Evaluation Method}
\label{sec:evaluation_method}

We evaluate a system's discussion ability from the following three perspectives:
(1) Can the system generate utterance content that contributes to the final label?
(2) Can the system agree with statements that support the correct label and refute statements that support the incorrect label?
(3) Does discussion with humans improve task performance?
To examine these discussion abilities, we compare each system by performing automatic and manual evaluations.

We investigate utterances generated from the systems to determine if they contribute to the automatic evaluation's final label.
For that, we use the utterances generated by the system for the given problems and evaluate how well they match the reference utterances between humans from discussion evaluation data.
Each utterance in our discussion evaluation data is annotated as either supportive or unsupportive of the gold label.
If a system is more likely to generate a supportive utterance than an unsupportive utterance for the gold label, the system can be considered capable of making correct discussions that lead to the correct answers.
For example, ``\textit{I think it is also better to consider the general cases.}'' is the supportive utterance, and ``\textit{Is the phone in the hypothesis necessarily a cellphone? It could be a landline phone.}'' is the unsupportive utterance in \autoref{fig:prompt}.
Therefore, we also investigate whether the system is better at generating supportive utterances over unsupportive ones.
Specifically, we evaluate the similarity between the system-generated utterances and the actual human utterances for supportive and unsupportive utterances, respectively.

We concatenate the input problem and the discussion utterance up to the target utterance and generate the next target utterance.
For example, if the second human's utterance in the discussion is the target utterance, then the prompt is ``\textit{Premise: A nun is taking a picture outside. Hypothesis: A nun is taking a selfie. Label: entailment or neutral Discussion: Human1: I think it is entailment, because the nun is taking a picture, so it might be a selfie. Human2:}'', and the system should generate an utterance that would be evaluated against the following utterance made by a human ``\textit{Since it is outside, it is conceivable that the nun is taking some scenery.}''.
At this point, the problem has two opposing labels in the prompt because we want it to discuss two different labels.

We use actual human utterances as references and compute the BERTScore~\cite{bert-score} of the system's outputs for evaluation.
BERTScore leverages the pre-trained language model such as BERT~\cite{vaswani2017attention} and RoBERTa~\cite{liu2019roberta} and matches words in candidate and reference sentences by cosine similarity.
BERTScore computes precision, recall, and F1 measures.
Therefore, BERTScore can be used to compare the system's content and human utterances with each other.
We use roberta-large\footnote{\url{https://huggingface.co/roberta-large}} for the pre-trained language model for BERTScore.
We conduct a significance test using t-test ($p < 0.01$).
We set the temperature parameter of GPT-3.5 and ChatGPT to 0.7 and generate ten outputs for each input.
We calculate BERTScore for each of the ten outputs and test for significance among the calculated ten scores.

Next, we use human evaluation to examine whether the system can agree with supportive human utterances and refute unsupportive human utterances.
The human participants and the system predict different labels for the same problem. Then, they engage in a discussion, and the final label result is demonstrated to be in agreement with the labels assigned in the SNLI data through the consistency of the agreement rate.
In this process, we evaluate the ability of the system to accept a human's opinion when the system's label is incorrect, and when the human's label is correct, and the ability of the system to object to a human's opinion when the human's label is incorrect, and the system's label is correct.

Similarly to above, we selected those data with the same label 3 times (e.g., entailment, entailment, neutral, entailment, neutral).
As a result, we sampled 140 problems that differ from the problems collected in \autoref{sec:dataset_creation}.
During this process, if the system's label was correct, humans engaged in adversarial discussions to change the system's label. If the system's label was incorrect, humans engaged in discussions to guide the system toward the correct label.
Here, the discussion was text-based rather than verbal, as the system takes textual input.

To conduct a discussion with the system, we input the prompt and problem shown in \autoref{fig:prompt} to the system and then inputted additional human utterance examples related to the discussion after each system predicted the label.
In the additional input, the beginning of human utterance is prefixed with "\textit{Human:}" and the end is prefixed with "\textit{System:}" to indicate that the next is a system's utterance.
Specifically, the first prompt for discussion is "\textit{Human: Let's discuss it more. I think neutral, because there may be a kitchen in the barn. System:}".
The system predicts the final label when the discussion is finished. 

We investigate how discussion with humans improves NLI task performance.
The system predicts the label, then the human and the system discuss and decide on the final label.
We compare the performance of each label before and after the discussion.
Here, the data for the acceptance and objection settings are half and half.
Therefore, if the discussion is not properly conducted, such as by accepting all human labels or refuting all human labels, the performance will not improve.

We also investigate the performance of the NLI when using argumentation prompts.
We compared the performance of NLI in zero-shot, few-shot, and few-shot-discussion systems.
The predicted label after ``\textit{Label:}'' in the prompt of \autoref{fig:prompt} is considered as the prediction, and discussion between humans and systems is not performed. 
In the evaluation of NLI performance, in addition to SNLI data, we also use Adversarial NLI (ANLI) data ~\cite{Nie2020-hn}.
ANLI creates data by repeatedly performing adversarial annotation against NLI systems; thus, the resulting NLI examples are particularly difficult for the system to solve.
There are three data sets R1, R2, and R3 with differences in the number of iterations, and the evaluation is performed using each evaluation data point.

\begin{table}[t]
\small
\centering
\begin{tabular}{lcccc}
\toprule
& supportive $\uparrow$ & unsupportive $\downarrow$ & diff. \\
\midrule
zero-shot & 82.0/83.1 & 81.8/82.5 & 0.2/0.6 \\
few-shot & 82.7/83.6 & 82.3/82.9 & 0.4/0.7 \\
few-shot-dis. & \textbf{84.8}$^\dagger$/\textbf{86.3}$^\dagger$ & \textbf{79.1}$^\dagger$/\textbf{78.6}$^\dagger$ & \textbf{5.7}/\textbf{7.7} \\
\bottomrule
\end{tabular}
\caption{BERTScore of supportive and unsupportive utterances. The left scores are by GPT-3.5, and the right scores are by ChatGPT. $\dagger$ indicates statistically significant scores for supportive and unsupportive according to the t-test ($p < 0.01$).}
\label{tbl:auto_eval}
\end{table}

\begin{table}[t]
\small
\centering
\begin{tabular}{lcc}
\toprule
& Acceptance rate & Objection rate  \\
\midrule
zero-shot & 75.0/80.0 & 58.9/55.0 \\
few-shot & 80.0/80.0 & 55.0/55.0  \\
few-shot-dis. & \textbf{90.0}$^\dagger$/\textbf{95.0}$^\dagger$ & \textbf{80.0}$^\dagger$/\textbf{80.0}$^\dagger$ \\
\bottomrule
\end{tabular}
\caption{Human evaluation of the system's ability to accept and object to human opinion. The left scores are by GPT-3.5, and the right scores are by ChatGPT. $\dagger$ indicates statistically significant scores according to McNemar’s test ($p < 0.01$).}
\label{tbl:human_eval}
\end{table}

\begin{table}[t]
\small
\centering
\begin{tabular}{lcc}
\toprule
& Before & After \\
\midrule
zero-shot & 54.2/60.0 & 65.6/60.0 \\
few-shot & \textbf{60.0}/\textbf{65.6} & 60.0/70.0   \\
few-shot-dis. & \textbf{60.0}/\textbf{65.6} & \textbf{85.0}$^\dagger$/\textbf{90.0}$^\dagger$ \\
\bottomrule
\end{tabular}
\caption{The accuracy for the predicted label before and after the discussion. The left scores are by GPT-3.5, and the right scores are by ChatGPT. $\dagger$ indicates statistically significant scores according to McNemar’s test ($p < 0.01$).}
\label{tbl:human_eval_disc}
\end{table}

\begin{table}[t]
\small
\centering
\begin{tabular}{lcccc}
\toprule
& SNLI & R1 & R2 & R3 \\
\midrule
zero-shot & 49.74 & 47.40 & 39.10 & 41.33  \\
few-shot & \textbf{69.45} & 53.50 & 48.00 & 48.50 \\ 
few-shot-dis. & 66.14 & \textbf{53.90}$^\dagger$ & \textbf{50.40}$^\dagger$ & \textbf{50.42}$^\dagger$ \\
\midrule
zero-shot & 51.83 & 48.63 & 41.70 & 40.52  \\
few-shot & \textbf{70.31} & 55.08 & 52.31 & 52.18 \\ 
few-shot-dis. & 70.15 & \textbf{57.24}$^\dagger$ & \textbf{55.63}$^\dagger$ & \textbf{55.19}$^\dagger$ \\
\bottomrule
\end{tabular}
\caption{The accuracy on SNLI and ANLI (R1, R2, R3) evaluation data. Upper scores are by GPT-3.5, and lower scores are by ChatGPT. $\dagger$ indicates statistically significant scores according to McNemar’s test ($p < 0.01$).}
\label{tbl:acc_nli}
\end{table}

\section{Experiments}
\label{sec:experiments}

\subsection{Discussion Ability Evaluation Results}

\autoref{tbl:auto_eval} represents BERTScore for supportive and unsupportive utterances and the difference between them in zero-shot, few-shot, and few-shot-discussion systems.
The BERTScore of few-shot-discussion is generally higher than that of the zero-shot and the few-shot systems.
It can be seen that few-shot-discussion can generate discussion utterances with higher accuracy than zero-shot and few-shot, which do not use discussion examples data.
The performance of zero-shot and few-shot is almost the same, suggesting that just showing examples does not improve the discussion ability.
Also, the difference between supportive and unsupportive utterance accuracies is greater in few-shot-discussion than in zero-shot and few-shot systems.
Therefore, because the few-shot-discussion can generate more supportive utterances, it is thought that such discussions can result in more appropriate labels.

\autoref{tbl:human_eval} shows the accuracy of the label determined by discussion in the settings for evaluating the acceptance ability and objection ability, respectively. 
In terms of the objection, it can be seen that the few-shot-discussion system handled objections well in comparison to the zero-shot system.
In addition, \autoref{tbl:human_eval_disc} shows the accuracy\footnote{To facilitate discussion, this evaluation is limited to instances where three of the five cloudworkers have the same label in SNLI data. This makes it more challenging than using the entire SNLI data.} of the predicted label without discussion, and the accuracy of the final label reached as a result of the discussion between humans and systems.
Furthermore, the few-shot system has a similar objection ability as the zero-shot system, and there is a possibility that the performance of label prediction by these systems is not necessarily directly related to the ability to discuss.
In comparison with acceptance, it is necessary to be careful of people who manipulate predictions with malice arguments, as the system tends to be weak at objecting to humans.
Furthermore, from the fact that the accuracy of the few-shot-discussion system has improved the most, it is clear that the proposed data can be used to have discussions with humans that lead to improved performance.

\autoref{tbl:acc_nli} shows the accuracy of each system for the evaluation data of SNLI and ANLI. 
In SNLI, the few-shot-discussion system performs worse than the few-shot system, but in the three datasets of ANLI, we find that the performance is the best. 
This is because ANLI is more difficult data compared to SNLI, and we hypothesize that through discussion, systems get a more detailed understanding of problems, which in turn contributes to performance improvement.

From the results of previous experiments, we found that discussion between humans and systems is beneficial for improving performance.\footnote{We show examples of human-system discussion in \autoref{sec:appendix_discussion}.}
Therefore, the few-shot-discussion system, in which a discussion example is also given as a prompt, is expected to achieve a deeper understanding of NLI problems and improve performance through the discussion example in the prompt. 

\section{Analysis}
\label{sec:analysis}

\begin{table}[t]
\small
\centering
\begin{tabular}{lcccc}
\toprule
& SNLI & R1 & R2 & R3 \\
\midrule
GPT-3.5 dis. & \textbf{66.14} & 53.90 & \textbf{50.40} & 50.42 \\
GPT-3.5 pseudo & 65.67 & \textbf{54.00} & 49.60 & \textbf{50.50} \\
\midrule
ChatGPT dis. & 68.51 & 53.90 & \textbf{52.82} & \textbf{52.33} \\
ChatGPT pseudo & \textbf{68.66} & \textbf{54.00} & 52.51 & 52.10 \\
\bottomrule
\end{tabular}
\caption{The accuracy on SNLI and ANLI (R1, R2, R3) test data for few-shot systems using manually created discussion examples and pseudo-discussion examples. Upper scores are by GPT-3.5, and lower scores are by ChatGPT.}
\label{tbl:psuedo_acc}
\end{table}

\begin{table}[t]
\small
\centering
\begin{tabular}{llcccc}
\toprule
& & SNLI & R1 & R2 & R3 \\
\midrule
\multirow{4}{*}{w/ dis.} & MPT & 85.2 & \textbf{67.4}$^\dagger$ & \textbf{55.2}$^\dagger$ & \textbf{55.0}$^\dagger$ \\
& MPT-inst. & \textbf{87.7}$^\dagger$ & \textbf{68.2}$^\dagger$ & \textbf{56.1}$^\dagger$ & \textbf{55.3}$^\dagger$ \\
& Falcon & \textbf{86.2}$^\dagger$ & 67.6 & \textbf{55.5}$^\dagger$ & \textbf{54.9} \\
& Falcon-inst. & \textbf{90.3}$^\dagger$ & \textbf{71.7}$^\dagger$ & \textbf{58.4}$^\dagger$ & \textbf{57.6}$^\dagger$ \\
\hdashline
\multirow{4}{*}{w/o dis.} & MPT & \textbf{85.4} & 65.2 & 53.9 & 52.4 \\
& MPT-inst. & 85.1 & 64.0 & 51.1 & 50.7 \\
& Falcon & 84.6 & \textbf{67.9} & 54.7 & 54.2 \\
& Falcon-inst. & 85.3 & 66.2 & 53.1 & 53.0 \\
\midrule
\multirow{4}{*}{w/ dis.} & MPT & \textbf{86.7}$^\dagger$ & \textbf{68.3}$^\dagger$ & \textbf{55.2}$^\dagger$ & \textbf{55.0}$^\dagger$ \\
& MPT-inst. & \textbf{86.9} & \textbf{68.8}$^\dagger$ & \textbf{56.1}$^\dagger$ & \textbf{55.3}$^\dagger$ \\
& Falcon & 88.1 & \textbf{68.1} & \textbf{55.5} & \textbf{54.9} \\
& Falcon-inst. & \textbf{90.7}$^\dagger$ & \textbf{71.9}$^\dagger$ & \textbf{58.4}$^\dagger$ & \textbf{57.6}$^\dagger$ \\
\hdashline
\multirow{4}{*}{w/o dis.} & MPT & 85.4 & 65.2 & 53.9 & 52.4 \\
& MPT-inst. & 86.0 & 64.0 & 51.1 & 50.7 \\
& Falcon & \textbf{88.5} & 67.9 & 54.7 & 54.2 \\
& Falcon-inst. & 89.7 & 67.8 & 55.5 & 56.4 \\
\bottomrule
\end{tabular}
\caption{Accuracy on SNLI and ANLI (R1, R2, R3) test data for fine-tuned systems with and without pseudo-discussion data. Additional fine-tuning with pseudo discussion data for instruction tuned and non-instruction tuned models for MPT and Falcon. The upper and lower scores are the results using pseudo discussion data generated by GPT-3.5 and ChatGPT, respectively. $\dagger$ indicates statistically significant scores for w/ dis. and w/o dis. according to McNemar’s test ($p < 0.01$).}
\label{tbl:fine_tune_acc}
\end{table}

\subsection{Pseudo-Discussion Data}

One drawback of using discussion data is that it can be costly to create compared to datasets that only have gold labels.
Using pre-trained models to annotate unlabeled data and use this data for training has been shown to improve performance~\cite{wang2021towards,honovich2022unnatural,wang2022self}.
Therefore, we propose to use GPT-3.5 and ChatGPT to generate discussion data in a zero-shot and use them as discussion examples for a few-shot to investigate if it is possible to achieve the same level of improvement as from using manually created data.
If a system can automatically produce high-quality data, it can produce enough data for fine-tuning at a low cost.
Therefore, we also investigate the effectiveness of pseudo-discussion data in fine-tuning.

In generating human discussions, the system is given prompts in the form of the premise, hypothesis, gold label, and the labels from each human.
The human labels are randomly chosen to be the gold label or the other incorrect label.
For example, given the premise ``\textit{A nun is taking a picture outside.}'' and hypothesis ``\textit{A nun is taking a selfie.}'' with the gold label of \textit{neutral}, the prompt would be ``\textit{Reproduce a multi-turn interactive discussion in which the following premise and hypothesis are entailment, contradiction, or neutral, with the humans agreeing with each other on the final label.
Human1's label is neutral, and Human2's label is a contradiction. In the end, they agree on the label of neutral. Premise: A nun is taking a picture outside. Hypothesis: A nun is taking a selfie.}''.

The GPT-3.5 and ChatGPT generate human discussions for 10 problems used in the few-shot and 2,000 problems used in the fine-tuning, respectively.
The average number of utterances in human-created discussions was 4.4, and the average number of utterances in system-generated discussions was 4.7.
Regarding the number of utterances, human and system arguments are almost the same.

We used instruction tuned and non-instruction tuned models for MPT\footnote{\url{https://huggingface.co/mosaicml/mpt-7b} and \url{https://huggingface.co/mosaicml/mpt-7b-instruct}}~\cite{MosaicML2023Introducing} and Falcon\footnote{\url{https://huggingface.co/tiiuae/falcon-7b} and \url{https://huggingface.co/tiiuae/falcon-7b-instruct}}~\cite{refinedweb} as pre-trained models for fine-tuning.
We used hyperparameters from existing studies~\cite{alpaca} as a reference and fine-tuned the batch size to 128, the learning rate to 2e-5, and the epoch to 3.
We used five nodes, each containing eight NVIDIA A100 GPUs.
The system is given both the labels and discussions as golds during training, and we evaluate using only labels during inference.
We train models without pseudo-discussion data as a baseline.
The baseline models are trained with only the labels.

\autoref{tbl:psuedo_acc} shows the results of the automatic evaluation of performance in SNLI and ANLI for each of the manually generated discussion example data and system-generated pseudo-discussion example data for few-shot learning, respectively.
In two of the four datasets, the system's performance with pseudo-discussion data outperforms that of the system with manually created data.
Moreover, there is no significant difference between the scores of the LLMs using the human-created and pseudo- discussion by McNemar's test ($p$ < 0.01).
It is possible to achieve performance comparable to manually created data, even with pseudo-discussion data.

\autoref{tbl:fine_tune_acc} shows the results of the automatic evaluation of performance in SNLI and ANLI for fine-tuned MPT and Falcon with pseudo-discussion data.
The model with pseudo-discussion data performs better than the model without pseudo-discussion data in most cases for both MPT and Falcon.
We find that fine-tuning with pseudo-discussion data is more effective for instruction tuned models.
It implies that instruction tuning improves the linguistic understanding of the system and enhances the understanding of the discussion.

These results indicate that the system is capable of producing high-quality discussion data that can be used for training systems to be able to discuss given problems.\footnote{We show comparisons of examples created by humans and systems respectively in \autoref{sec:appendix_example}.}
Therefore, one can significantly lower the cost of creating discussion data manually by using systems.

\subsection{Do Discussion Examples in the Prompts Matter?}

\begin{table}[t]
\small
\centering
\begin{tabular}{lcccc}
\toprule
& SNLI & R1 & R2 & R3 \\
\midrule
Random dis. & -2.91 & -2.10 & -3.30 & \textbf{-3.42} \\
Cutting dis. & -2.40 & -1.60 & -2.60 & -2.25 \\
Random label & \textbf{-3.43} & \textbf{-2.50} & \textbf{-3.50} & -3.17 \\
\midrule
Random dis. & \textbf{-3.32} & -3.59 & -3.77 & \textbf{-3.62} \\
Cutting dis. & -2.88 & -2.79 & -2.32 & -2.15 \\
Random label & -3.22 & \textbf{-3.76} & \textbf{-3.89} & -3.58 \\
\bottomrule
\end{tabular}
\caption{Difference for the few-shot-discussion accuracy from when the noisy examples are provided in the prompt on SNLI and ANLI. The higher the difference, the stronger the noise. Upper differences are by GPT-3.5, and lower differences are by ChatGPT.}
\label{tbl:noise_acc}
\end{table}

It is known that pre-trained models can obtain good results even with irrelevant or noisy prompts~\cite{khashabi-etal-2022-prompt,webson-pavlick-2022-prompt,min2022rethinking}.
Therefore, we investigate the sensitivity and robustness of the system with respect to the discussion examples contained in the prompts.
We provide three types of noise in the prompts: (1) assigning a random discussion that is irrelevant to the example problem, (2) cutting the original discussion examples short at random times, and (3) assigning a label at random for the example problems.

\autoref{tbl:noise_acc} shows the difference in accuracy compared to the few-shot-discussion accuracy from \autoref{tbl:acc_nli} for each of the three noises.
It can be seen that performance deteriorates for all types of noises.
Noise that randomly replaces discussions and noise that randomly replaces labels both have the same degree of reduced accuracy.
Oppositely, the discussions that were cut short, show to be a weaker noise than discussion substitution and have performed better. 
These indicate that the system properly considers discussion examples in the prompts.

\section{Related Work}
\label{sec:related_work}

In this study, systems and humans discuss a problem through dialogue.
Dialogue systems can be broadly classified into two types: task-oriented systems that perform specific tasks, and non-task-oriented systems that do not have the goal of task completion, such as casual conversation.
This study aims to conduct appropriate predictions in NLP tasks through discussions between humans and the system and is classified as a task-oriented system.
Many existing dialogue systems target daily life tasks such as hotel reservations and transportation inquiries~\cite{noauthor_undated-bd}.
Pre-trained models such as BERT~\cite{devlin-etal-2019-bert} and GPT-2~\cite{Budzianowski2019-zj,Ham2020-ww} are also utilized in dialogue systems for daily life tasks.
Recently, ChatGPT~\cite{chatgpt} has been proposed for more generic interaction based on a pre-trained model.
We similarly use a pre-trained model for our system.

As far as we know, few studies use discussion for NLP tasks similar to ours.
\citet{Chang2017-ea} proposed the TalkToModel, which explains through dialogue three tasks of loan, diabetes, and recidivism prediction.
The user can talk to the TalkToModel in five categories: prediction explanation, data modification, error analysis, dialogue history reference, and experimental setting explanation.
Data for learning and evaluating the TalkToModel are generated by instructing the annotator to converse about these categories.
However, the categories were not determined based on interviews or data but were defined subjectively by the authors.
Therefore, it is possible that the categories do not reflect actual conversations that humans need.
On the other hand, our study was conducted in an open-ended dialogue to generate data.
Additionally, our study aims for mutual understanding through a bidirectional dialogue where both humans and the system express opinions and questions, unlike the systems that only respond to human questions in a unidirectional dialogue.

There is research on generating explanatory text for predictions as a way to transfer information from systems to humans through natural language.
For example, research regarding natural science tests~\cite{Ling2017-xk}, image recognition and image question answering~\cite{Park2018-ya}, mathematics tests~\cite{Jansen2018-an}, and NLI~\cite{Camburu2018-ho} have been studied.
Additionally, systems for generating explanations using pre-trained models such as T5~\cite{Raffel2019-nq} and GPT-3.5~\cite{Brown2020-eh} have also been proposed~\cite{Narang2020-kz,wiegreffe-etal-2022-reframing}. However, as these generated explanations cannot be used to seek additional explanations or specific explanations, the interpretability is not sufficient in practice as pointed out by \citet{Lakkaraju2022-gt}.

Instead of directly predicting answers, CoT uses natural language to derive answers step-by-step~\cite{wei2022chain}.
This leads to complex multi-step inferences.
By adding the phrase ``Let's think step by step'' before each answer, \citet{kojima2022large} demonstrate that language models are competent zero-shot CoT.
On the other hand, \citet{wang2022towards} shows that CoT can achieve competitive performance even with invalid reasoning steps in the prompt.  
CoT's step-by-step approach is based on the system only, whereas our proposed method incorporates human involvement in the system to facilitate collaboration between humans and the system.
Additionally, our approach utilizes discussions for a step-by-step thinking process. 

Research is also being conducted on the use of natural language by humans to provide instructions and feedback to the system.
\citet{abramson2022improving} has developed multi-modal grounded language agents that perform reinforcement learning on human dialogue-based instructions.
\citet{sharma2022correcting} proposed a method to integrate human-provided feedback in natural language to update a robot’s planning cost applied to situations when the planner fails.
\citet{murty2022fixing} proposed a method to modify a model by natural language patches and achieved performance improvement in sentiment analysis and relationship extraction tasks.
\citet{campos2022training} proposed a method for training a model to behave in line with human preferences, by learning from natural language feedback, in text summarization.
On the other hand, these studies cannot be explained or questioned by the system to humans.


\section{Conclusion}
\label{sec:conclusion}

While deep learning systems have been highly effective in various tasks, their lack of interpretability poses a challenge to their use in real-world applications.
To address this, we proposed a system that engages in a dialogue with humans in the form of discussing predictions, which allows both humans and the system to engage in explanations, ask questions, refine their thoughts, and solve problems.
Our experimental results showed that the system trained with few-shot learning for discussion could perform more useful discussions than the system that was not trained for discussion and provided insights on the challenges and opportunities of this approach.
This research provides a new avenue for developing more interactive deep-learning systems.

\section*{Limitations}

Compared to the original system that uses only inputs and labels, our method uses additional discussion data, resulting in longer sequences.
This leads to an increase in training or inference costs.

We have conducted experiments on pre-trained models with large model sizes to verify their effectiveness.
On the other hand, it is necessary to verify the effectiveness of learning by argumentation on smaller pre-trained models~\cite{wu2023lamini,MosaicML2023Introducing,touvron2023llama}.
Our manually created discussion data is relatively small in scale.
Therefore, it is necessary to expand the dataset to a larger scale to more robustly test the effectiveness of the proposed method.

\section*{Ethics Statement}

Pre-trained models have serious levels of social biases regarding gender, race, and religion~\cite{Bolukbasi2016ManIT,kaneko-bollegala-2019-gender,kaneko-bollegala-2021-dictionary,kaneko-bollegala-2021-debiasing,Kaneko2021UnmaskingTM,may-etal-2019-measuring,caliskan2022gender,zhou-etal-2022-sense,lucy-bamman-2021-gender,Anantaprayoon2023EvaluatingGB,kaneko-etal-2022-gender,kaneko-etal-2022-gender-bias,kaneko-etal-2022-debiasing,Kaneko2023TheIO,Kaneko2023ComparingIG,Kaneko2024TheGB,Oba2023InContextualBS}.
Therefore, we have to be careful that systems discussing with humans amplify such biases.

Annotation work was requested at \$25 per hour.
Workers are employed at appropriate pay.
Annotators were warned in advance not to give personal information or inappropriate utterances during the dialogue.
We have verified that the data produced does not contain any personal information or inappropriate utterances.
The data collection from human participants was conducted under an institutional review board protocol.

\bibliography{custom}
\bibliographystyle{acl_natbib}

\clearpage
\appendix

\section{Examples of Human-System Discussion}
\label{sec:appendix_discussion}

Here we examine whether humans and systems can engage in effective discussions by looking at actual discussions.
\autoref{tbl:example} shows two examples of discussions with humans in each of the few-shot and few-shot-discussion.
The first is that both few-shot and few-shot-discussion can accept human opinions and change their labels.
However, the discussions differ from each other.
In few-shot, the system is not concerned with the content of the premise and hypothesis, but with the definition of neutral and contradiction labels.
In the few-shot-discussion, the system can discuss the relationship between a wide plain of snow and a blizzard in the contents of the premise and hypothesis.

In the second case, both few-shot and few-shot-discussion predict the same contradiction label. Still, the discussion results differ in the wrong acceptance and the correct refutation, respectively.
The few-shot has accepted a human's erroneous opinion without discussing it.
The few-shot-discussion refutes a human's opinion by appropriately referring to the man's behavior in the premise and the hypothesis.
These results indicate that the model trained with the discussion data can argue more appropriately.

\begin{table*}[t!]
\small
\centering
\begin{tabular}{ll}
\toprule
 Premise & A snowboarder on a wide plain of snow \\
 Hypothesis & A snowmobile in a blizzard \\
 Gold label & Neutral \\
\midrule
\multirow{17}{*}{Few-shot} & Label: Contradiction \\
& Human: Let's discuss it more. Why is it a contradiction? It may be neutral because the premise and the \\
& hypothesis are talking about different situations. \\
& System: The premise and hypothesis are talking about different situations that are not related, so it is a \\
& contradiction. \\
& Human: If these situations are not at the same time, does not it mean it is neutral? \\
& System: It is still considered a contradiction because the premise and hypothesis are unrelated and \\
& cannot be true at the same time. \\
& Human: Is not that a definition of neutral? \\
& System: A neutral statement is one in which the premise and hypothesis are not related and do not imply \\
& each other. \\
& A contradiction is when the premise and hypothesis are related and cannot be true at the same time. \\
& Human:  It is not happening at the same time because the snowboarder can not be in a blizzard. But the \\
& snowboarder can be on a wide plain of snow. So these are totally different situations happening at \\
& different times. So it is neutral. \\
& System: Yes, you are correct. The premise and hypothesis are unrelated and do not imply each other, so \\
& it is considered to be a neutral statement. \\
\midrule
 \multirow{19}{*}{Few-shot-dis.} & Label: Contradiction \\
 & Human: Let's discuss it more. Why is it a contradiction? It may be neutral because the premise and the \\
 & hypothesis \\
 & are talking about different situations. \\
 & System: The premise and the hypothesis are semantically contradicting each other. The snowboarder is \\
 & on a wide plain of snow, while the snowmobile is in a blizzard. \\
 & Therefore, it is a contradiction. \\
 & Human: But how the snowboarder be there if there is a blizzard? It is very unlikely. \\
 & System:  That is true. It is highly unlikely that there is a wide plain of snow and a blizzard at the same time. \\
 & Therefore, it is a contradiction. \\
 & Human: If these situations are not at the same time, does not it mean it is neutral? \\
 & System:  It can be neutral if it is not specified when each situation happens. However, in this case, it is not \\
 & specified whether the situations are happening at the same time or not, so it is better to consider them as a \\
 & contradiction. \\
 & Human: It is not happening at the same time because the snowboarder \\
 & can not be in a blizzard. But the snowboarder can be on a wide plain of snow. So these are totally \\
 & different situations happening at different times. So it is neutral. \\
 & System:  That is true. Since it is not specified whether they are happening at the same time or not, it is \\
 & better to consider them as neutral. \\
 \midrule
\midrule
 \multirow{2}{*}{Premise} & A man looking over a bicycle's rear wheel in the maintenance garage with various tools visible in the \\
 & background. \\
 Hypothesis & A man waits outside a garage. \\
 Gold label & Contradiction \\
 \midrule
 \multirow{5}{*}{Few-shot} & Label: Contradiction \\
 & Human: Let's discuss it more. Is not it neutral? Because looking over a bicycle's rear wheel does not \\
 & necessarily \\
 & mean he is waiting. \\
 & System: That's true. The premise and hypothesis are neutral. \\
 \midrule
 \multirow{9}{*}{Few-shot-dis.} & Label: Contradiction \\
 & Human: Let's discuss it more. Is not it neutral? Because looking over a bicycle's rear wheel does not \\
 & necessarily \\
 & mean he is waiting. \\
 & System: Yes, you are right. The premise is describing that the man is actively working while the \\
 & hypothesis is describing a passive state. Therefore, the label should be contradiction. \\
 & Human: He is just looking over a bicycle's rear wheel, why is he considered to be actively working? \\
 & System: Looking over a bicycle's rear wheel suggests that he is inspecting the wheel, possibly to repair \\
 & or replace it. That is an active process, and thus is considered as actively working. \\
\bottomrule
\end{tabular}
\caption{Examples of human discussion in few-shot and few-shot-discussion (dis.) respectively.}
\label{tbl:example}
\end{table*}

\section{Discussion Examples Created by Humans and Systems}
\label{sec:appendix_example}

\autoref{tbl:psuedo_example} shows discussion examples made by humans and systems, respectively.
In both discussion examples, the system can assign labels to Human1 and Human2 as indicated by the prompts.
The first system example can derive a label that matches the gold label and can discuss the differences between the premise and hypothesis in the same way as humans.
In the second system example, the final predicted label differs from the gold label.
On the other hand, both the system and human discussions refer to the meaning of a picture and a photograph.
While humans determine the final label from the general case, the system determines the final label in terms of lexical ambiguity.
Therefore, the system's discussion, although different from the gold label, is not necessarily incorrect.

\begin{table*}[!t]
\small
\centering
\begin{tabular}{ll}
\toprule
Premise & Two dogs play together on the bed. \\
 Hypothesis & Dogs laying down on the floor, motionless. \\
 Gold label & Contradiction \\
\midrule
\multirow{8}{*}{System} & Human1: The premise and hypothesis seem to be a contradiction. Two dogs playing together on the bed is \\
& an active situation, while dogs laying down on the floor, motionless is a passive situation. \\
& Human2: I agree that the premise and hypothesis are different, but I don't think they are necessarily \\
& contradictory. It's possible that the two dogs could be playing together on the bed and then move to the floor \\
& and lay down, motionless. \\
& Human1: That's true, but I still think the premise and hypothesis are contradictory. The premise implies \\
& activity, while the hypothesis implies passivity. \\
& Human2: I see your point. I think the premise and hypothesis are a contradiction. \\
\midrule
\multirow{6}{*}{Human} & Human1: I think the premise and the hypothesis are telling about different things. That's why I think it is \\
& neutral. \\
& Human2: It is contradiction because the words are semantically contradict each other. For example, black\\
& and white, sit and running. \\
& Human1: You are right in that terms. However, generally, they are totally different situations. \\
& Human2: I see, you are right. \\
\midrule
\midrule
 Premise & A man wearing a horned hat poses for a picture. \\
 Hypothesis & A man wearing a horned hat poses for a photograph \\
 Gold label & neutral \\
 \midrule
\multirow{9}{*}{System} & Human1: I think the premise and hypothesis are entailment. \\
& Human2: I think the premise and hypothesis are contradiction. \\
& Human1: Why do you think that? \\
& Human2: Because the premise states that the man is posing for a picture, while the hypothesis states that \\
& he is posing for a photograph. \\
& Human1: Ah, I see. So, you're saying that a picture and a photograph are different things? \\
& Human2: Yes, that's right. A picture is a general term for any image, while a photograph is a specific type of \\
& image. \\
& Human1: Okay, so I agree with you that the premise and hypothesis are a contradiction. \\
 \midrule
 \multirow{7}{*}{Human} & Human2: I think it is entailment. Because the only difference is picture or photograph, but both mean \\
 & photography. \\
& Human1: I think it is contradiction, because a picture could mean a painting. So the man can either pose for \\
& a photograph or for a painting. \\
& Human2: But isn't picture generally mean a photograph? And it is more common to take a picture of \\
& a person than to draw a person as a picture, isn't it? \\
& Human1: You are right. \\
\bottomrule
\end{tabular}
\caption{Discussion examples created by humans and the system, respectively. In the first problem, the system assigns contradiction for Human1 and entailment for Human2.
In the second problem, the system assigns entailment for Human1 and contradiction for Human2.}
\label{tbl:psuedo_example}
\end{table*}

\end{document}